\documentclass{article}

\usepackage{natbib}
\setcitestyle{numbers,square,comma}
\usepackage[preprint]{neurips_2025}
\usepackage{graphicx}
\usepackage{paralist}
\usepackage{amsmath}
\usepackage{wrapfig}
\usepackage{multirow}
\usepackage{multicol}
\usepackage{paralist}
\usepackage{color}
\usepackage[table]{xcolor}

\usepackage[utf8]{inputenc} 
\usepackage[T1]{fontenc}    
\usepackage[pagebackref=true,breaklinks=true,colorlinks,bookmarks=false,citecolor=blue,linkcolor=blue]{hyperref}
\usepackage{url}            
\usepackage{booktabs}       
\usepackage{amsfonts}       
\usepackage{nicefrac}       
\usepackage{microtype}      
\usepackage{xcolor}         
\usepackage{enumitem}

\title{CoCo4D: Comprehensive and Complex 4D Scene Generation}
\newcommand{\name}{CoCo4D}

\author{Junwei Zhou$^1$\quad Xueting Li$^2$\quad Lu Qi$^{3,4,*}$\quad Ming-Hsuan Yang$^{5}$\quad \\
\vspace{-0.5em} \\
$^1$Huazhong University of Science and Technology \quad
$^2$NVIDIA \quad \\
$^3$Wuhan University \quad
$^4$Insta360 Research \quad
$^5$UC Merced
}

\begin{document}
\renewcommand{\thefootnote}{}
\footnotetext{$^*$Corresponding author}

\maketitle

\vspace{-.2in}
\begin{figure}[h]
    \centering
    \includegraphics[width=1.0\linewidth]{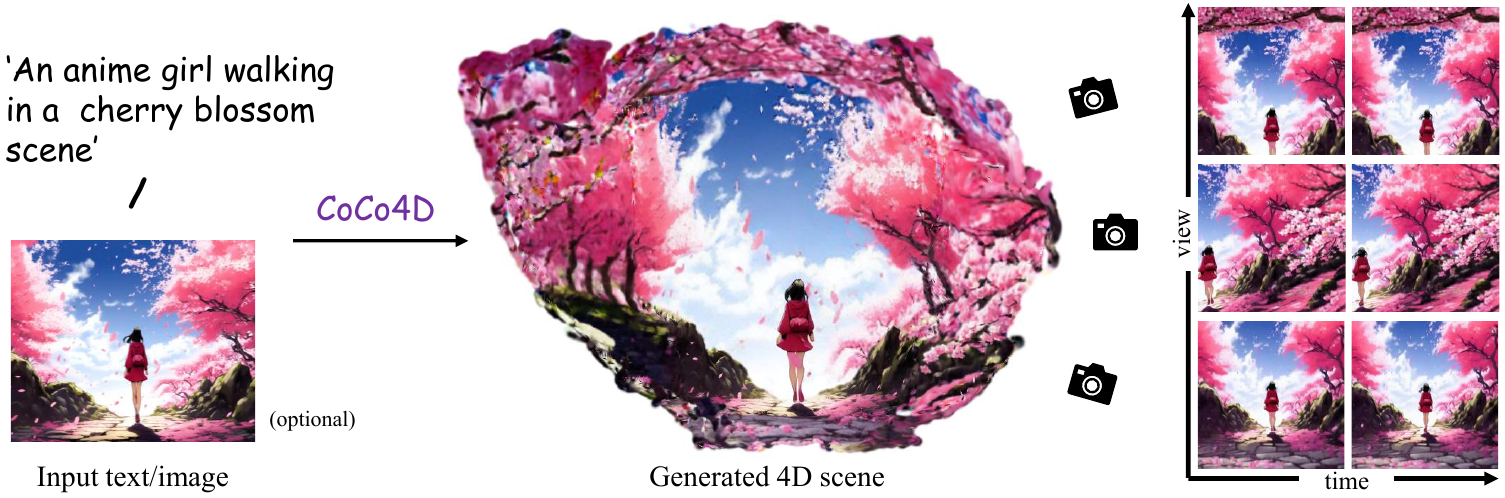}
    \vspace{-.3in}
    \caption{\name~generates a comprehensive and complex 4D scene with input conditions, achieving high-quality and consistent multi-view rendering at each timestamp.}
    \label{fig:teaser}
    \vspace{-.05in}
\end{figure}


\begin{abstract}
    \vspace{-.1in}
    Existing 4D synthesis methods primarily focus on object-level generation or dynamic scene synthesis with limited novel views, restricting their ability to generate multi-view consistent and immersive dynamic 4D scenes.
    To address these constraints, we propose a framework (dubbed as \name) for generating detailed dynamic 4D scenes from text prompts, with the option to include images. Our method leverages the crucial observation that articulated motion typically characterizes foreground objects, whereas background alterations are less pronounced. Consequently, \name~divides 4D scene synthesis into two responsibilities: modeling the dynamic foreground and creating the evolving background, both directed by a reference motion sequence.    
    Given a text prompt and an optional reference image, \name~first generates an initial motion sequence utilizing video diffusion models. This motion sequence then guides the synthesis of both the dynamic foreground object and the background using a novel progressive outpainting scheme. To ensure seamless integration of the moving foreground object within the dynamic background, \name~optimizes a parametric trajectory for the foreground, resulting in realistic and coherent blending.
    Extensive experiments show that 
    \name~achieves comparable or superior performance in 4D scene generation compared to existing methods, demonstrating its effectiveness and efficiency.
    More results are presented on our website \url{https://colezwhy.github.io/coco4d/}.
\end{abstract}
\vspace{-8pt}

\section{Introduction}
\label{sec:intro}
Creating 4D dynamic scenes is a fundamental challenge in computer vision and graphics. 
While recent rapid development in diffusion models has advanced object-level motion generation~\citep{xie2024sv4d,bahmani20244dfy}, synthesizing fully immersive and complex 4D scenes with dynamic foreground and background remains largely underexplored.

Existing 4D scene generation methods primarily utilize score distillation-based optimization or novel view synthesis-based learning. 
Score distillation methods~\cite{singer2023text, bahmani20244dfy} typically optimize a 4D scene representation (e.g., DynamicNeRF~\cite{Gao-ICCV-DynNeRF}) by exploiting prior in diffusion models~\cite{chen2023videocrafter1,khachatryan2023text2video} through various Score Distillation Sampling (SDS) strategies~\cite{poole2022dreamfusion}. 
Though they generate multi-view consistent results, they suffer from impractical optimization time (e.g., 24 hours per scene) and lack ambient dynamics. 
Furthermore, these methods mainly focus on dynamic foreground modeling while assuming no or over-simplified background motion, which also hinders the application of these methods.
Conversely, learning-based methods employ training multi-view video diffusion models~\cite{bahmani2024ac3d} or novel view synthesis techniques to generate novel spatial and temporal views~\cite{gupta2024paintscene4d} from a single or multi-view images, and then optimize a 4D representation using the multi-view videos as supervision. These approaches are more efficient compared to optimization-based methods and enable richer background motion thanks to the novel view synthesis techniques. However, training video diffusion models requires massive datasets, and the training-free variants~\cite{liu2025free4d, lee2024vividdreamgenerating3dscene} struggle with multi-view consistency across wide camera ranges.

In this work, we generate comprehensive (wide view range) and complex (dynamic foreground and background with sophisticated motion) 4D scenes within an acceptable time. Our key insight lies in the distinct motion patterns of scene components: foreground objects exhibit rapid articulations (e.g., jumping animals), while backgrounds evolve gradually (e.g., swaying trees). Based on this intuition, we design \name, a general framework that takes a text prompt and an optional single view image as input to generate a 4D scene with an extrapolated dynamic background and foreground. We start by generating an initial motion (i.e., a reference video sequence) based on the input. To capture the rapid motion of the foreground object, we design a pipeline that reconstructs the 3D foreground object and animates it under the guidance of the reference video. To synthesize an extrapolated and evolving background, we propose a novel progressive outpaint strategy that gradually expands the camera range of the background scene by inpaint-project-optimize loops, leveraging the prior of video inpainting models~\cite{zi2024cococo}. Finally, to align the foreground object and background, we learn a trajectory for the foreground object and compose it with the background using depth knowledge. As a result, \name~can generate a comprehensive and complex 4D scene under one hour, which is significantly faster than existing methods that take 4$\sim$24 hours.
We evaluate \name~against state-of-the-art 4D scene reconstruction and generation methods on a validation dataset. The main contributions of this work are:
\begin{itemize}[leftmargin=*, labelsep=0.5em]
\vspace{-2mm}
    \item We propose \name, a practical approach for highly comprehensive and complex 4D scene generation, demonstrating superior performance compared to state-of-the-art methods.
    \vspace{-1mm}
    \item We design different pipelines tailored to generate dynamic foreground and background based on their motion characteristics, which are then unified by learning a plausible trajectory and depth-aware composition.
      \vspace{-1mm}
    \item Extensive experiments demonstrate that \name~achieves superior qualitative and quantitative results compared to baselines with low time cost. 
\end{itemize}
\vspace{-5pt}

\section{Related Work}
\label{sec:related}
\vspace{-.1in}
\subsection{4D Object Generation}
\vspace{-.1in}
Recently, with the rapid advancement in the field of 3D object generation~\cite{li2024craftsman, xiang2024structured, zhou2024layout}, the current research has increasingly shifted towards the domain of 4D object generation, aiming to push the boundaries of generative vision by incorporating temporal dynamics.
As the first attempt, Animate124~\citep{zhao2023animate124} generates
4D objects from images through the score distillation sampling (SDS) from both multi-view~\citep{shi2023mvdream} and video diffusion models~\citep{chen2023videocrafter1}.
After that, various works focus on improving the fidelity of 4D generation under a simplified video-to-4D setting.
For example, Consistent4D~\citep{jiang2024consistentd} produces 360° dynamic objects using SDS loss in combination with a post-video enhancement module.
SC4D~\citep{wu2024sc4d} introduces sparse controlled points to model motion in 4D space.
SV4D~\citep{xie2024sv4d}, 4Diffusion~\citep{zhang20244diffusion}, and Animate3D~\citep{jiang2024animate3d} train multi-view video diffusion models to provide temporally-coherent novel view synthesis. 
Additionally, L4GM~\citep{ren2024l4gm} takes a different route by extending feed-forward 3D models into the 4D domain, enabling a more efficient generation pipeline, albeit with reduced performance.
Inspired by those works, we consider our foreground branch as a form of 4D object generation, capturing both spatial structure and temporal dynamics.

\vspace{-5pt}
\subsection{Video-to-4D Reconstruction}
\vspace{-3pt}
Monocular video-to-4D reconstruction has recently emerged as a promising research direction, considering that videos can provide more spatial information with each frame.
Early efforts rely on optimization-based methods, which have achieved promising results.
Shape-of-motion~\citep{som2024} leverages depth and tracking priors to optimize a 4D scene with an input monocular video.
DreamScene4D~\citep{dreamscene4d} reconstructs dynamic foreground objects and refines the motion through motion factorization.
Monst3R~\citep{zhang2024monst3r} fine-tunes the stereo-based Dust3R~\citep{dust3r_cvpr24} to reconstruct spatial geometry and temporal motion jointly.
More recent works~\citep{lu2024align3r,liang2025feedforwardbullettimereconstructiondynamic,lei2024mosca} have followed this trend that train feed-forward models for monocular video-to-4D reconstruction. 
In our work, we adopt a reconstruction-like strategy by first reconstructing the 4D scene at a reference view, followed by a novel extrapolation process that progressively completes the scene across space and time.

\vspace{-5pt}
\subsection{4D Scene Generation}
Generating a complicated 4D scene with realistic motion and multi-view consistency remains challenging.
An early milestone in this direction is the work of Singer et al.~\citep{singer2023text}, which
generates a fully dynamic 4D scene with a 360° visible range.
Specifically, it introduces a hybrid SDS strategy to distill motion patterns from existing video diffusion models.
Then, several works propose novel score distillation techniques to improve the generated 4D scene quality.
For example, 4D-fy~\citep{bahmani20244dfy} uses hash encoding to separate the static and dynamic representations,
while Dream-in-4D~\citep{zheng2024unified} adopts a two-stage strategy that first generates the static asset and then infuses motion using video-based SDS.
TC4D~\citep{bahmani2024tc4d} further augments 4D objects with sequential movements using a trajectory-aware SDS.
Trans4D~\citep{zeng2024trans4d} models intrinsic scene transitions with a specially designed network.
Despite these advancements, most methods struggle to construct complicated and coherent motions, resulting in inconsistent dynamics and high computational costs that degrade overall visual quality.
%
A series of works~\citep{yu20244real, gupta2024paintscene4d, lee2024vividdreamgenerating3dscene, liu2025free4d} take advantage of the strong properties of 4D Gaussian Splatting (4DGS)~\cite{4dgs_Wu_2024_CVPR}, collecting the multi-view videos to train the 4DGS representation.
However, due to the inherent drawbacks of current video diffusion models~\citep{blattmann2023stable, yu2024viewcrafter}, they
still face difficulties in producing wide field-of-view scenes with temporally consistent motion
In our work, we adopt a reconstruction-inspired strategy for generating the background scene, aiming to achieve high-quality motion and spatial consistency while maintaining scalability.

\vspace{-.05in}
\section{Method}
\label{sec:method}
\vspace{-.08in}
%
%
%

We present CoCo4D, a novel framework that generates a comprehensive and complex 4D scene from a text prompt $y$, with an optionally single-view image $I$.
%
%
%
%

\begin{figure}[ht]
    \centering
    \includegraphics[width=1.0\linewidth]{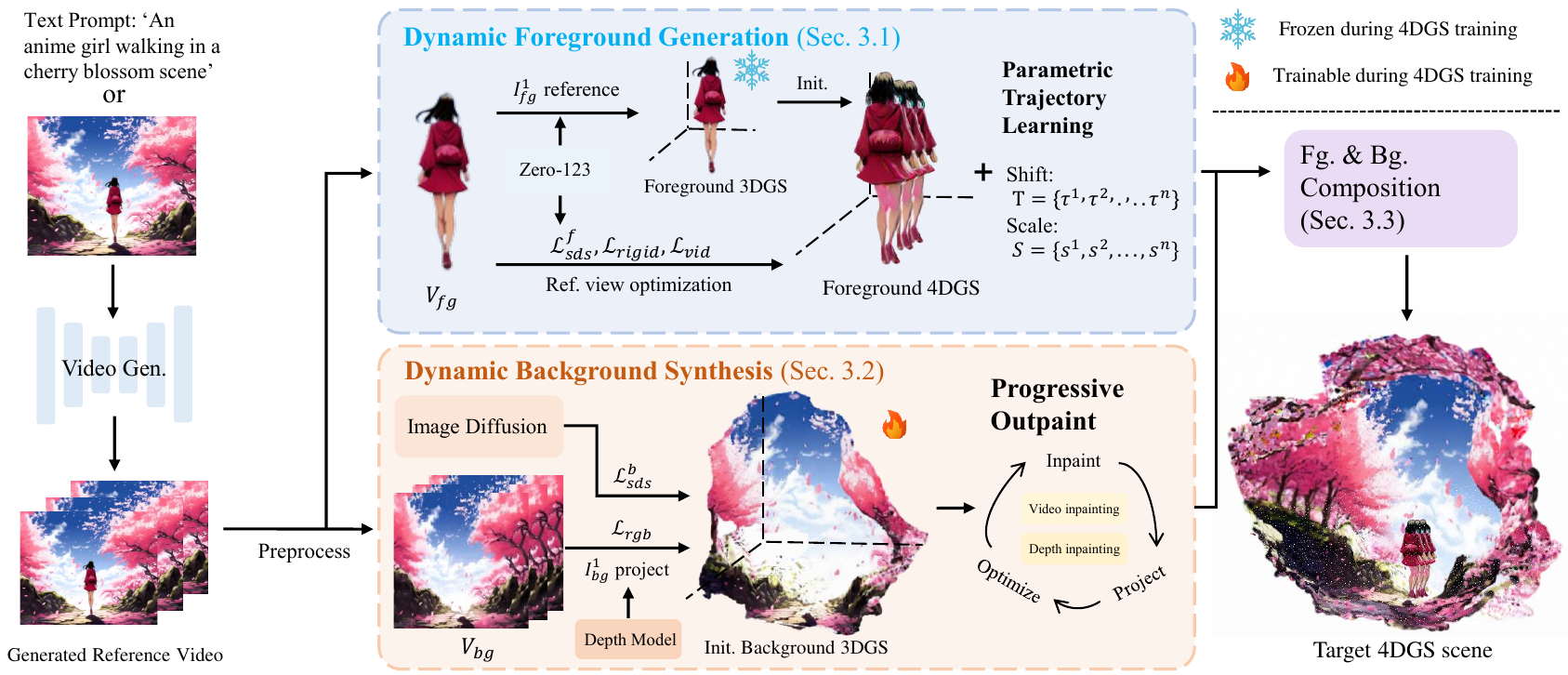}
    \vspace{-.2in}
    \caption{Overview of \name. We first generate the reference video (initial motion) for both foreground and background, and disentangle them for separate generation, i.e., Dynamic Foreground Generation in Sec.~\ref{subsec:fgbranch} and Dynamic Background Synthesis in Sec.~\ref{subsec:bgbranch}. Finally, we precisely compose these two components in Sec.~\ref{subsec:scenecomp} to form our comprehensive and complex target 4D scene.}
    \label{fig:pipeline}
    \vspace{-.2in}
\end{figure}

We begin by creating an initial reference video based on the input conditions.
If both an image and a text prompt are given, we leverage an image-to-video (I2V) model~\cite{xing2023dynamicrafter} to synthesize the reference video. 
If only a text prompt is available, we either employ a text-to-video (T2V) model~\cite{yang2024cogvideox}, or first generate an image using diffusion models before creating the video with an I2V model. 
We denote the resulting reference video including $n$ frames as $V_{ref} = \{I^1_{ref}, I^2_{ref}, ..., I^n_{ref}\}$.
To separate the foreground object from the background, we employ SAM2~\citep{ravi2024sam2} to decompose the reference video into foreground frames $V_{fg}=\{I^1_{fg}, I^2_{fg}, ...,I^n_{fg}\}$, background frames $V_{bg}=\{I^1_{bg}, I^2_{bg}, ...,I^n_{bg}\}$, and foreground masks $M_{fg}=\{m^1, m^2, ...,m^n\}$. 
%
%
%
%
%
%
For articulated foreground object synthesis, we reconstruct a static 3D model of the object and learn its motion dynamics supervised by $V_{fg}$, augmented with physics-based constraints (Sec.~\ref{subsec:fgbranch}). 
The dynamic background generation begins with projecting the first frame of the reference video to the 3D space using depth information, followed by scene extrapolation and animation through our novel progressive outpaint technique (Sec.~\ref{subsec:bgbranch}).
Finally, we assemble the final scene by deriving an optimal trajectory for the foreground object and performing depth-aware composition with the animated background (Sec.~\ref{subsec:scenecomp}).
An overview of CoCo4D is shown in Fig.~\ref{fig:pipeline}.

\subsection{Dynamic Foreground Generation}
\label{subsec:fgbranch}
\vspace{-.08in}
Our foreground generation module aims to generate an animated 3D object sequence that closely matches the appearance and motion described in the reference video generated by the text prompt.

\vspace{-8pt}
\paragraph{Static 3D Foreground Object Reconstruction.} 
A straightforward approach~\cite{xie2024sv4d} is to utilize a pre-trained video diffusion model to synthesize multi-view temporal frames for dynamic 3D sequence reconstruction. While computationally efficient, these methods struggle to predict spatially consistent frames, resulting in collapsed geometry.
Instead, we adopt an optimization-based manner that first reconstructs the foreground 3D object, followed by a motion optimization stage, both guided by the reference video.
We start by uniformly sampling a set of 3D Gaussians, $\mathcal{G}_f$, on the surface of a sphere. To align with the object depicted in the first frame $I^1_{fg}$ of the reference video, we render the Gaussians from the corresponding camera view $c^1$ (referred to as the ``seen view'') and compute the L2 loss by:
\begin{align}
    \mathcal{L}^f_{rgb} = ||I^1_{fg} - \mathcal{R}(\mathcal{G}_f, c^1)||_2.
    \label{eq2:foreground_rgb_loss}
\end{align}
where $\mathcal{R}$ denotes the rendering function of Gaussian Splatting~\cite{kerbl3Dgaussians}.
We further hallucinate the unseen views by leveraging a multi-view diffusion model (i.e., Zero123~\cite{liu2023zero}) via Score Distillation~\cite{tang2023dreamgaussian}. Specifically, we render 3D Gaussians $\mathcal{G}_f$ from uniformly sampled camera poses $c$ and compute the Score Distillation Sampling (SDS) loss by:
\begin{align}
    \nabla_{\mathcal{G}_f} \mathcal{L}^f_{sds} = \mathbb{E}_{\gamma,\epsilon,c}[w(\gamma)(\hat{\epsilon}(x_f; c, I^1_{fg}, \gamma) - \epsilon)\frac{\partial x_f}{\partial \mathcal{G}_f}],
    \label{eq1:sds}
\end{align}
where $x_f=\mathcal{R}(\mathcal{G}_f, c)$, $\hat{\epsilon}$ is the score function that predicts the sampled noise $\epsilon$ from rendered image $x_f$ with noise-level $\gamma$, $I^1_{fg}$ is the first frame that serves as condition input to Zero123, and $w(\gamma)$ is the weighting function~\cite{song2022denoisingdiffusionimplicitmodels}.
\vspace{-8pt}
\paragraph{Foreground Object Motion Optimization.}
After optimizing the static 3D object using both $\mathcal{L}^f_{rgb}$ and $\mathcal{L}^f_{sds}$, we animate the foreground object by learning a deformation network $D_f$ ~\cite{ren2023dreamgaussian4d}. The deformed Gaussians at each timestamp $t$ is denoted as $\mathcal{G}_f^t=D_f(\mathcal{G}_f, t).$
To preserve the learned geometry and texture, we freeze parameters of the static 3D Gaussians $\mathcal{G}_f$ throughout the motion learning process.
%
Since the reference video provides strong supervision for object motion, we render deformed Gaussians at each timestamp from the seen view and denote the rendered reference view frames as {\small$\overline{V}_{ref}=\{\overline{I^1_{fg}}, \overline{I^2_{fg}}, ..., \overline{I^n_{fg}}\}$}, where {\small$\overline{I^t_{fg}}=\mathcal{R}(\mathcal{G}^t_f,c^1)$}.
The reconstruction loss can be computed by:
\begin{align}
    \mathcal{L}_{vid} = ||I^t_{fg} - \overline{I^t_{fg}}||_2.
\label{eq3:vidrgbloss}
\end{align}
For unseen views at each timestamp $t$, we similarly leverage Zero123 via Score Distillation (Eq.~\ref{eq1:sds}) to ensure plausible geometry during motion learning, using {\small$I^t_{fg}$} as the conditional image.
%
To enhance motion plausibility, we implement a rigidity loss $\mathcal{L}_{rigid}$~\citep{luiten2024dynamic} that limits excessive displacement between neighboring Gaussians, preserving motion fidelity during optimization.
The overall loss for training the 4D foreground at timestamp $t$ can be written as:
\begin{align}
\mathcal{L}_{4dfg}=\mathcal{L}^f_{sds}+\mathcal{L}_{vid}+\mathcal{L}_{rigid}.
\label{eq4:4dfg}
\end{align}

\paragraph{Foreground Object Trajectory learning.}
During the foreground object reconstruction process described above, we translate and scale the object at each timestamp to the image centers to facilitate foreground trajectory. And to restore the object to its original position in the video frame at each timestamp, we apply a shift ($\tau^t=\{\mu^t_x, \mu^t_y\}$) and scale $s^t$. Thus, the trajectory of an object can be represented as $T=\{\tau^1,\tau^2, ..., \tau^n\}$ and $S=\{s^1, s^2, ..., s^n\}$.
We further refine the trajectory parameters $S$ and $T$ by optimizing them with the RGB video alignment loss (Eq.~\ref{eq3:vidrgbloss}), ensuring the object trajectory aligns precisely with the reference video. 
More details are illustrated in the Appendix.~\ref{app:implementation}.

\subsection{Dynamic Background Synthesis}
\label{subsec:bgbranch}
\vspace{-5pt}
Different from the articulated foreground object, the background scene typically exhibits gradual, slow-moving changes. In the following, we first introduce how to initialize a 3D background scene from the reference video, and then detail the processes of background expansion and animation using a novel progressive outpaint scheme.

\vspace{-7pt}
\paragraph{3D Background Scene Initialization.}
We begin by projecting the background depicted in the first frame $I_{bg}^1$ of the reference video into the 3D space using an estimated depth map {\small$\mathcal{D}_{bg}^1=\mathcal{DM}(I_{bg}^1)$}. 
This creates a point cloud ({\small$\mathcal{PC}=\{p^1, p^2, ...,p^{HW}\}$}) including $H\times W$ points, where $H, W$ are the height and width of the video frame. Each point in $\mathcal{PC}$ corresponds to a pixel in the first frame {\small$I_{bg}^1$}. For every point in $\mathcal{PC}$, we initialize a background Gaussian using the point's spatial coordinates and corresponding RGB values as the initial Gaussian position and color attributes.
All background Gaussians collectively form the 3D background scene representation $\mathcal{G}_b$.
This simple 2D-to-3D lifting process results in low-quality 3D Gaussians and holes caused by occlusions. To improve scene quality, we refine the Gaussians through optimization using both the reference frame and Score Distillation Sampling.
Specifically, we render $\mathcal{G}_b$ at the seen view $c^1$ and compute the L2 loss between the rendering and reference frame as:
\begin{align}
    \mathcal{L}^b_{rgb}=||\mathcal{R}(\mathcal{G}_b,c^1)-I^1_{bg}||_2.
    \label{eq5:bgrgb}
\end{align}
We further enhance the quality of this reference view by applying the SDS loss:
\begin{align}
    \nabla_{\mathcal{G}_b} \mathcal{L}^b_{sds}(x_b) = \mathbb{E}_{\gamma^*,\epsilon}[w(\gamma^*)(\hat{\epsilon}(x_b; I^1_{bg}, \gamma^*) - \epsilon)\frac{\partial x_b}{\partial \mathcal{G}_b}],
    \label{eq6:bgsds}
\end{align}
here $x_b=\mathcal{R}(\mathcal{G}_b,c^1)$ is the rendered background image at the seen view $c^1$. After optimization, we obtain a high-quality static 3D background with plausible reference view. Next, we describe how to expand and animate this static 3D scene using progressive outpaint.

\vspace{-10pt}
\paragraph{Progressive Outpaint.}
%
%
%
To simultaneously expand and animate the background scene, we propose an \textcolor{red}{inpaint}-\textcolor{green}{project}-\textcolor{blue}{optimize} loop.
In the first iteration, we start by equipping the optimized static 3D background scene $\mathcal{G}_b$ with a deformation network $D_b$ and then optimize $D_b$ and $\mathcal{G}_b$ using losses from~\citep{4dgs_Wu_2024_CVPR}: a L1 reconstruction loss $\mathcal{L}_1$ and a total-variation regularization~\cite{cao2023hexplane} $\mathcal{L}_{tv}$.
To expand the initial background scene, we define $k_1$ camera poses ({\small$C^1_{out}=\{c^1_1, c^1_2, ..., c^1_{k1}\}$}) that look at empty background regions that we require our method to outpaint. The superscript 1 of {\small$C^1_{out}$} indicates that these cameras are used in the first iteration for the inpaint-project-optimize loop.
For each camera pose $c^1_r$ in $C^1_{out}$, we render deformed 3D Gaussians at each timestamp to obtain a video sequence  ({\small$V^{c^1_r}=\{x^{c^1_r}_1, x^{c^1_r}_2, ...x^{c^1_r}_n\}$}) and a mask sequence ({\small$M^{c^1_r}=\{m^{c^1_r}_1,m^{c^1_r}_2,...,m^{c^1_r}_n \}$}) at camera pose $c^1_r$, i.e., {\small$x^{c^1_r}_t, m^{c^1_r}_t = \mathcal{R}(D_b(\mathcal{G}_b, t), c^1_r)$}.
Then we \textcolor{red}{inpaint} the rendered videos to fill the empty part in the background.
Naively inpainting these rendered images using video inpainting models leads to inconsistent results.
To resolve this, we propose using image inpainting as guidance. Specifically, we inpaint the first frame of the rendered video sequence with Stable Diffusion and the input text prompt, i.e., {\footnotesize$\widetilde{x}^{c^1_r}_1=Inpaint(x^{c^1_r}_1, m^{c^1_r}_1, y)$}.
We then copy and paste the inpainted region into each subsequent frame in {\footnotesize$V^{c^1_r}$} to form a pseudo video     {\footnotesize$\widetilde{V}^{c^1_r}=\{\widetilde{x}^{c^1_r}_1, \widetilde{x}^{c^1_r}_2, ..., \widetilde{x}^{c^1_r}_n\}$}, each $\widetilde{x}^{c^1_r}_1$ represents the frame after ``copy and paste''.
Finally, we add a small amount of noise to each frame's ``copy and pasted'' region and refine these frames by a video inpainting model~\cite{zi2024cococo} $\mathcal{VI}$.
The pseudo video generation and video inpainting can be formally formulated as:
%
\begin{align}
    \widetilde{x}^{c^1_r}_t=Concat({x}^{c^1_r}_t, \widetilde{x}^{c^1_r}_1 \odot m^{c^1_r}_t), \quad V^{c^1_r}_{bg}=\mathcal{VI}(\widetilde{V}^{c^1_r}, M^{c^1_r}, strength),
    \label{eq7:pseudo_vid}
\end{align}
where {\footnotesize$V^{c^1_r}_{bg}$} is the final video at camera pose $c^1_r$ for the later training of 4D Gaussians, and $strength$ is the lowered noise level.
By utilizing the inpainted region from the first frame as guidance, our model produces more consistent video inpainting results, as shown in Fig.~\ref{fig:pseudo_vid}.

To convert the inpainted video content from the 2D space to the 3D scene, we \textcolor{green}{project} the inpainted pixels of the first frame in the video {\scriptsize $V^{c^1_r}_{bg}$} to extend the 3D Gaussians using depth information as how we initialize the point cloud.
In order to seamlessly integrate these new Gaussians into the 3D scene, we apply a depth inpainting model that fills the depth map rendered from {$\mathcal{G}_b$} at camera pose $c^1_r$, followed by projecting new Gaussians to the 3D scene using the inpainted depth.
Note that we slightly misuse the notation and still denote the 3D Gaussians after such integration as $\mathcal{G}_b$.
\begin{wrapfigure}{r}{0.58\textwidth}
\vspace{-.15in}
  \begin{centering}
\includegraphics[width=0.6\textwidth]{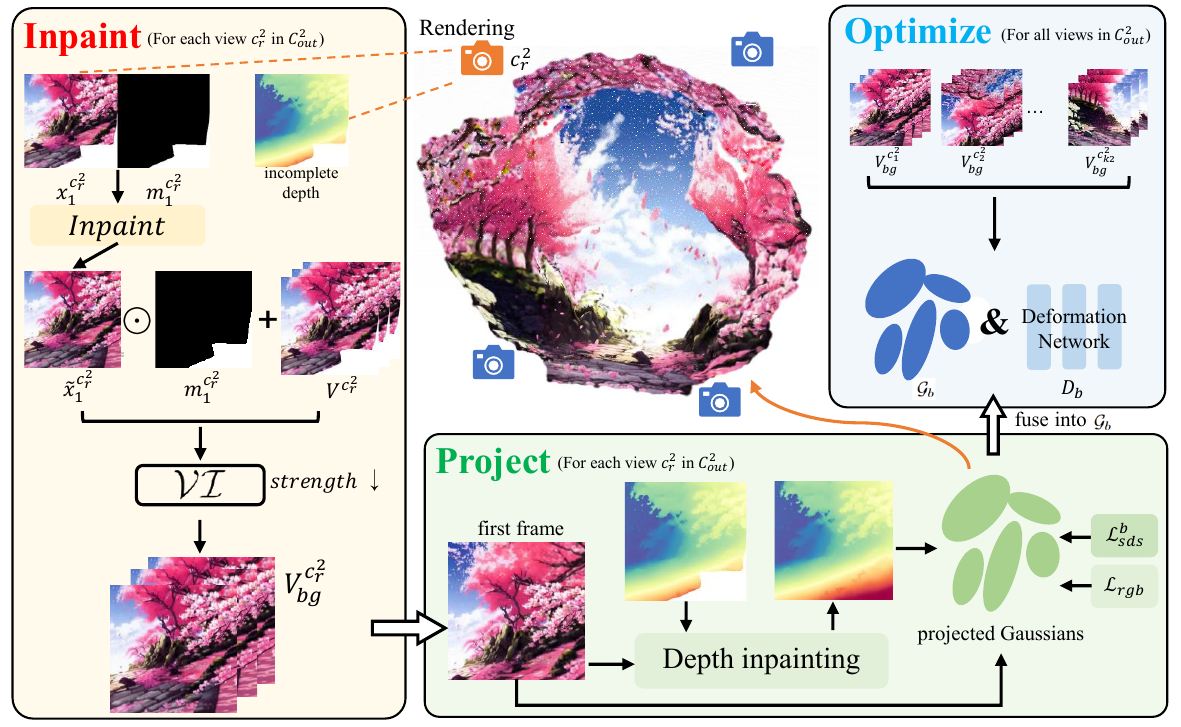}
  \end{centering}
  \vspace{-.2in}
  \caption{A simple illustration of the second round of the inpaint-project-optimize loop.}
  \label{fig:outpaint}
  \vspace{-.3in}
\end{wrapfigure}            
Finally, we refine the fused static background Gaussians $\mathcal{G}_b$ with the L1 and SDS loss described in Eq.~\ref{eq5:bgrgb} and Eq.~\ref{eq6:bgsds} by changing the reference image to {\scriptsize$\widetilde{x}^{c^1_r}_1$}.

Next, we \textcolor{cyan}{optimize} the deformation network together with Gaussians using the reference video and curated novel view frames {\scriptsize$\{V^{c^1_1}_{bg}, V^{c^1_2}_{bg},...,V^{c^1_{k1}}_{bg}\}$}, which facilitate the animation of 3D Gaussians $\mathcal{G}_b$.

Through experimentation, we observe that expanding the background scene by a large area in a single step leads to inferior results. Thus, we can carry out multiple inpaint-project-optimize iterations, incrementally expanding the scene by a small amount each time.
Given the 4D Gaussians $\mathcal{G}_b$ and $D_b$ produced by the first iteration,
We set another $k_2$ camera poses $C^2_{out}$ and repeat the inpaint, project, and optimize process to further expand the background.
A detailed illustration of the second iteration is shown in Fig.~\ref{fig:outpaint}.
After a few rounds of such iterations, we obtain a dynamic 3D background with consistent motion that covers a large camera range, as shown in Fig.~\ref{fig:teaser}.
More details are presented in Appendix.~\ref{app:implementation}.

\subsection{Foreground and Background Composition}
\label{subsec:scenecomp}
Given the optimized dynamic foreground object and background scene, we combine them seamlessly by exploiting depth information.
We use the same depth estimation model $\mathcal{DM}$ as in Sec.~\ref{subsec:bgbranch} to predict the depth map {\footnotesize$\mathcal{D}_{ref}=\mathcal{DM}(I^1_{ref})$} for the first frame of the reference video, which contains both the foreground and background.
This composition process can be seen as an affine transformation that warps the foreground object into the coordinate system of the background scene.
We render the depth from both foreground and background Gaussians at each timestamp from the reference view with default camera intrinsics, denoted as $\mathcal{D}^f$ and $\mathcal{D}^b$.
Using the background as a reference, we calculate the distance from the camera center to the foreground position in the background coordinate system as the depth shift:
{\footnotesize
\begin{align}
    d^f = (mean(\mathcal{D}_{ref}\odot m^1) - \min(\mathcal{D}_{ref}))*\Delta + \min(\mathcal{D}^b),
    \label{eq8:depth}
\end{align}}
here $\Delta$ is the relative scaling factor of value ranges between the background and foreground:
{\footnotesize
\begin{align}
    \Delta =  \frac{\max(\mathcal{D}^b \odot \sim m^1) - \min(\mathcal{D}^b \odot \sim m^1)}{\max(\mathcal{D}_{ref} \odot \sim m^1) - \min(\mathcal{D}_{ref} \odot \sim m^1)}.
    \label{eq9:relscale}
\end{align}}
Since we use the same depth model $\mathcal{DM}$ as in background synthesis, the value ranges of two different depth maps are well aligned, which provides an accurate estimation of the foreground depth in the background coordinate system.
In addition, we use another scaling factor $\varepsilon$ to indicate the scale change of the foreground content when composing the scene.
To compute this scaling factor, we render the foreground object together with the background and obtain the mask $m_f$ of the foreground object. We then compute the scale change $\varepsilon$ as a simple width division between the bounding boxes of $m_f$ and the ground truth mask in $M_{fg}$.
%
%
After that, we accordingly adjust the foreground Gaussians' screen space parameters with $\varepsilon$ to achieve the trajectory after composing.
For the screen space scales $S$, we simply multiply each $s^t$ with $\varepsilon$.
And for screen space shift, we scale the interval between each two neighboring $\tau^{t-1}$ and $\tau^t$ by:
{\footnotesize
\begin{equation}
\hat\tau^t = \left\{
\begin{aligned}
&\tau^t, & t = 1 \\
&\varepsilon \cdot (\tau^2 - \tau^1) + \hat\tau^{t-1}, & t > 1 
\end{aligned}
\right.
\label{eq10:traj}
\end{equation}}

Finally, we compose the foreground and background 4D Gaussian representation during rendering, which results in the comprehensive and complex target 4D scene.

\section{Experiments}
\label{sec:exp}
\vspace{-5pt}
\subsection{Implementation Details}
\label{subsec:imple}
We use DynamiCrafter~\citep{xing2023dynamicrafter} and CogVideoX~\citep{yang2024cogvideox} as our I2V and T2V model discussed in Sec.~\ref{sec:method}.
In dynamic foreground generation, we place the poses with a camera radius of 2.5 and an azimuth interval of 10 degrees around the Gaussians. We train the 3DGS representation with 1000 iterations and 4DGS representation with 800 iterations in a batch size of 4. 
The foreground Gaussians are frozen during training of the deformation network.
In the background branch, we use the depth inpainting model from ~\cite{engstler2024invisible}, and the Stable Diffusion v2.1 for the SDS loss $\mathcal{L}^b_{sds}$ in Eq.~\ref{eq6:bgsds} and SDXL~\cite{podell2023sdxlimprovinglatentdiffusion} for pseudo video generation.
The noise ratio for $\mathcal{L}^b_{sds}$ is set to 0.5 by default.
We adopt COCOCO~\cite{zi2024cococo} as our video inpainting model, and the $strength$ is set to 0.7 to obtain consistent and ambient dynamics.
For progressive outpaint, we repeat our inpaint-project-optimize loop 2 times, collecting videos from 9 camera poses as training data.
Additional experimental configurations and training hyper-parameters can be found in Appendix~\ref{app:implementation}.

\subsection{Main Results}
\label{subsec:main_results}
In this section, we provide quantitative results and qualitative results to demonstrate the superiority of \name{} in generating a fully dynamic 4D scene with complicated motion patterns.
For methods~\cite{gupta2024paintscene4d, lee2024vividdreamgenerating3dscene,liu2025free4d} that synthesize 4D scenes via generating multi-view video frames, we compare with Free4D~\cite{liu2025free4d}, which provides the open-source implementation.
We also compare with score distillation-based methods, i.e., Dream-in-4D~\cite{zheng2024unified} and 4D-fy~\cite{bahmani20244dfy}, under the setting of text-to-4D scene generation.
We conduct experiments with our self-captured text prompts and input reference images (optional).
Since no standardized benchmark exists for this task, we curated a set of text prompts (with optional image inputs) to evaluate our method and baseline approaches.

\begin{table}[ht]
\centering
\small
\vspace{-3pt}
\renewcommand\arraystretch{1.2}
{\footnotesize \caption{Quantitative results of \name and other 4D scene generation methods. We conduct comparisons on both VBench metrics and CLIP-Score. The best and second best results are \textbf{bold} and \underline{underlined}, respectively.}}
\vspace{-5pt}
\resizebox{\linewidth}{!}{
\begin{tabular}{l|ccccc}
\toprule
Method & Time cost $\downarrow$  & CLIP Score & Motion Smoothness  & Dynamic Degree  & Aesthetic Quality \\
\midrule
4D-fy  &  24 hours & 22.32\% & 79.62\% & 43.29\% & 52.98\%\\ 
Dream-in-4D  &  13.5 hours & \textbf{23.58\%} & \underline{89.84\%} & 56.32\% & 53.31\% \\ 
Free4D  &  \textasciitilde 1 hour & 22.06\% & 82.18\% & \underline{66.39\%} & \underline{60.97\%}\\ 
\midrule
\name    &  \textasciitilde 1 hour & \underline{23.47\%} & \textbf{92.37}\% & \textbf{70.95\%} & \textbf{63.86}\%\\ 

\bottomrule
\end{tabular}
}
\label{tab:vbench}
\vspace{-.1in}
\end{table}

\begin{table}[ht]
\small
\centering
\vspace{-5pt}
\caption{User study on different 4D scene generation methods in terms of motion quality, multi-view consistency, and visual quality. The best and second-best results are shown in \textbf{bold} and \underline{underlined}.}
\begin{tabular}{l | ccc}
\toprule
Method & Motion Quality  & Multi-view Consistency & Visual Quality \\
\midrule
4D-fy    & 5.68 & \textbf{7.82} & 4.43  \\ 
Dream-in-4D    &  \underline{6.19} & \underline{7.45} & 5.56 \\  
Free4D    &  5.42 & 6.97 & \underline{7.63} \\ 
\midrule
Ours    &  \textbf{7.04} & 7.26 & \textbf{7.84} \\ 
\bottomrule
\end{tabular}
\label{tab:user}
\vspace{-0.1in}
\end{table}

\vspace{-.1in}
\paragraph{Quantitative Results.}
For fair comparisons, we use the Vbench~\citep{zhang2024evaluationagent} metrics to evaluate all the methods.
We adopt Motion Smoothness, Dynamic Degree, and Aesthetic Quality to evaluate the overall quality of rendered videos.
Moreover, we adopt CLIP Score~\cite{park2021benchmark} to evaluate whether the synthesized videos match the given text prompt.
We first present the video quality assessment results in Tab.~\ref{tab:vbench}, together with the time cost to demonstrate the efficiency of all methods.
As the table demonstrates, our proposed \name{} shows comparable or superior performance compared with existing 4D scene generation methods across different metrics, especially the ones with articulated motions and obvious trajectory.
Next, we carry out a user study and show the results in Tab.~\ref{tab:user}.
Using this study, we evaluate three key aspects of the 4D scene rendered videos: motion quality, multi-view consistency, and visual quality.
30 users were surveyed to rate the generated 4D scenes using these three metrics. 
%
%
Overall, it is obvious that \name{} outperforms other compared methods in appearance and motion quality, which demonstrates \name{}'s capability in generating complex motion and realistic 4DGS scenes.

\begin{figure}[t]
    \centering
    \includegraphics[width=1.0\linewidth]{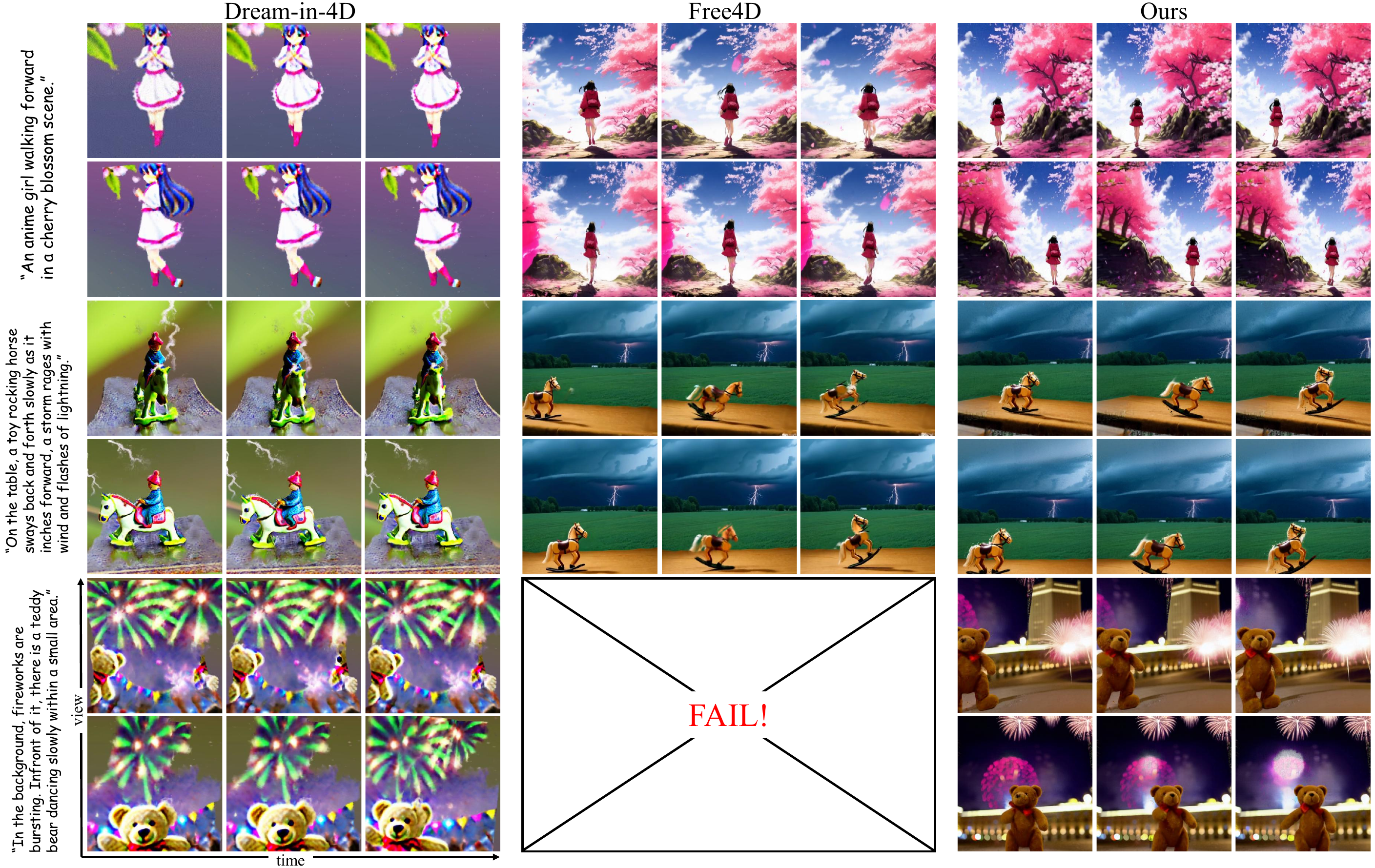}
    \vspace{-.22in}
    \caption{Qualitative comparison of our method against other 4D scene generation methods.}
    \label{fig:qualitative_main}
    \vspace{-.12in}
\end{figure}

\begin{figure}[h!]
    \centering
    \includegraphics[width=1.0\linewidth]{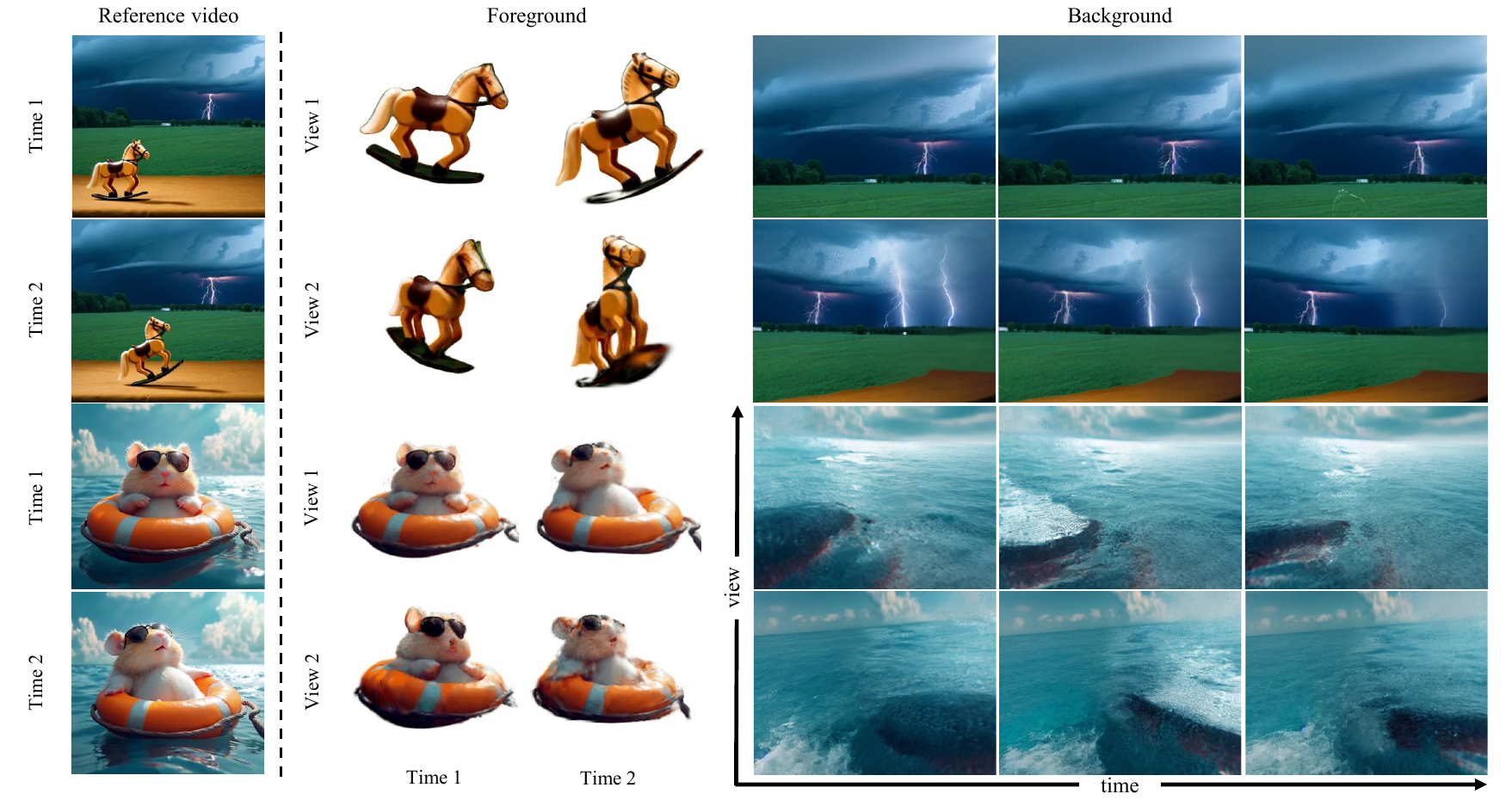}
    \vspace{-.2in}
    \caption{Examples of separated foreground and background 4DGS rendered views in two scenes.}
    \label{fig:seperate}
    \vspace{-.15in}
\end{figure}

\begin{figure}[htbp]
\begin{minipage}[c]{0.6\linewidth}
    \centering
\includegraphics[width=1.\linewidth]{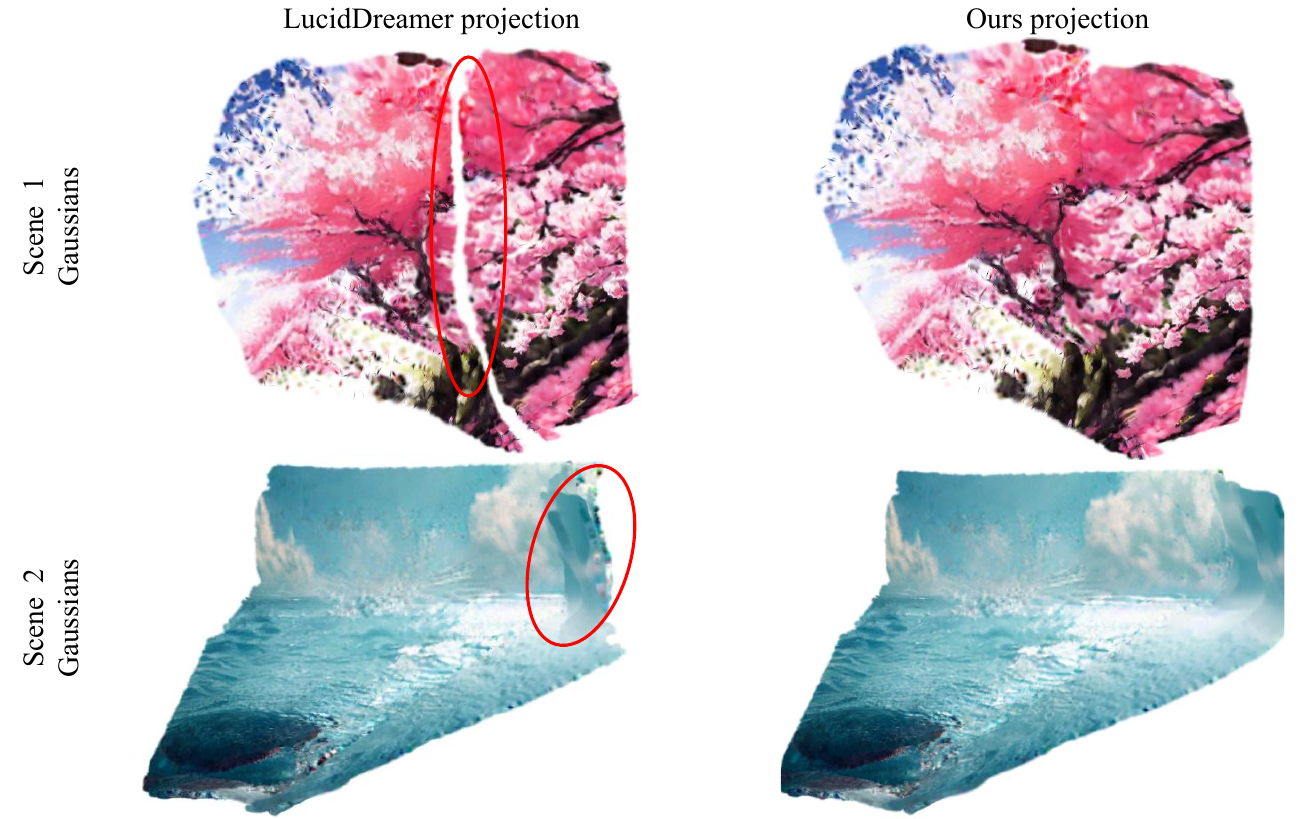}
\vspace{-16pt}
    \caption{ Ablation studies on the effectiveness of our depth inpainting model guided projection operation.}
    \protect\label{fig:depthinpaint}
\end{minipage}
\hfill
\begin{minipage}[c]{0.38\linewidth}
   \centering
  \includegraphics[width=1.\linewidth]{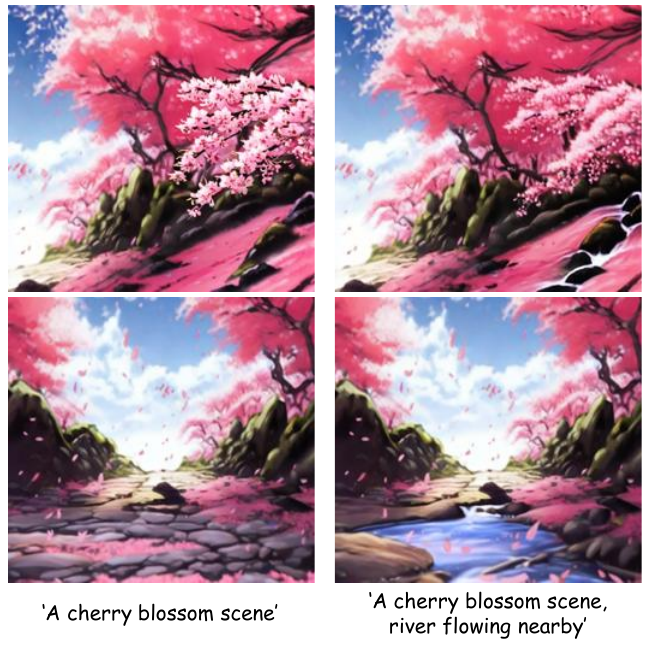}
\vspace{-16pt}
    \caption {An example of customized background 4D scene outpaint.}
    \protect\label{fig:cus}
\end{minipage}
\vspace{-.1in}
\end{figure}

\vspace{-20pt}
\paragraph{Qualitative Results.}
We compare \name{} with Dream-in-4D and Free4D, then present the visualization results in Fig.~\ref{fig:qualitative_main}.
We can see that though score-distillation methods show good multi-view consistency, they struggle to produce sophisticated motion.
Moreover, Free4D occasionally fails due to the unsuccessful multi-view video generation and stereo initialization,
while \name{} enables complex motion across a wide camera range.
In addition, we visualize the foreground and background separately in Fig.~\ref{fig:seperate}.
We also present an example of 4DGS outpaint with customized text prompt $y$ in Fig.~\ref{fig:cus}, showing that we can use the text prompt to indicate the desired outpaint content, then generate an ideal 4D scene as an added benefit.
More visualization results and a detailed analysis of the metrics are present in Appendix.~\ref{app:vis} and Appendix.~\ref{app:analy}, respectively.

\begin{figure}[h!]
    \centering
    \includegraphics[width=1.0\linewidth]{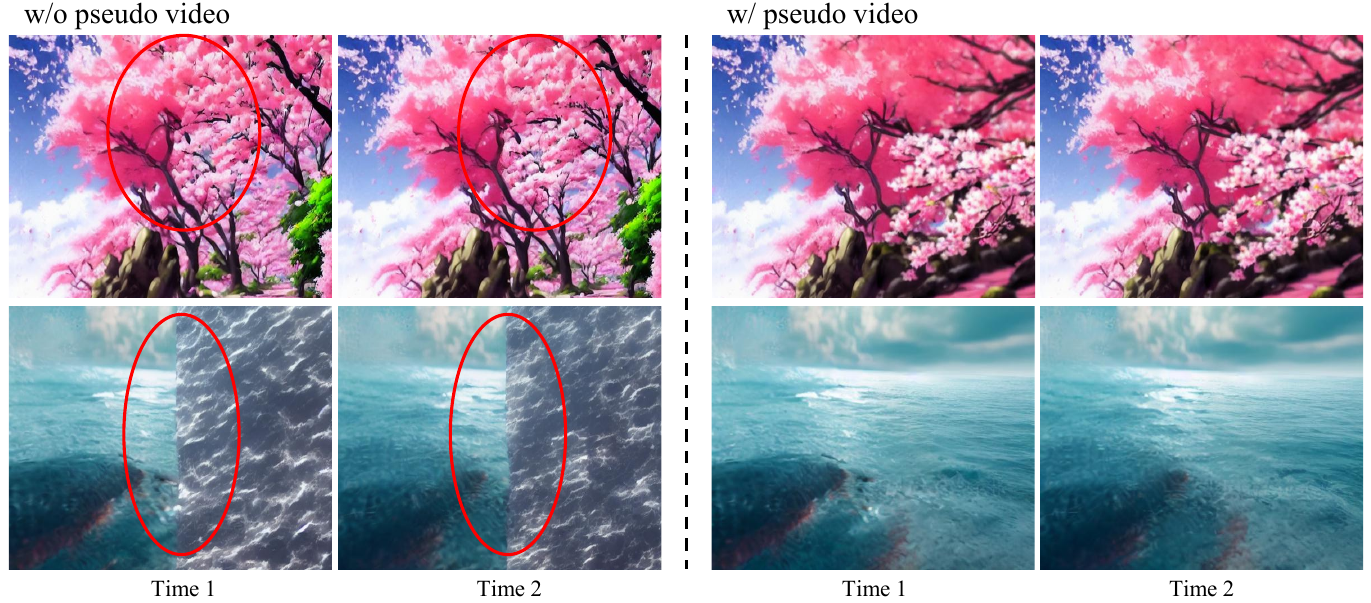}
    \vspace{-.2in}
    \caption{Ablation study on the effectiveness of our pseudo video generation strategy. We show two background scenes for example.}
    \label{fig:pseudo_vid}
    \vspace{-5pt}
\end{figure}

\vspace{-5pt}
\subsection{Ablation Studies}
\vspace{-5pt}
\paragraph{Pseudo Video Generation.} We compare video inpainting results with and without our image inpainting guided strategy.
As illustrated in Fig.~\ref{fig:pseudo_vid}, video inpainting models alone often generate inconsistent results, while our guided inpainting strategy produces seamless inpainting together with reasonable motion, which enhances the quality of the extended part of our generated 4DGS scene.

\textbf{Depth Inpainting.} We compare our 3D Gaussians projection operation with the one in LucidDreamer~\cite{chung2023luciddreamer} in Fig.~\ref{fig:depthinpaint}. It is clear that with the incorporation of depth inpainting model, our projection can achieve seamless point cloud fusion, while the one in LucidDreamer can not.

\textbf{Training losses.} We analyze the effect of different losses in the progressive outpaint. First, as we mentioned in Sec.~\ref{subsec:bgbranch}, the SDS loss $\mathcal{L}^b_{sds}$ will help enhance the quality of different views, and equip the Gaussians with adequate ability in novel view synthesis, which helps fill holes in the current scene and outpaint to another view.
Second, after projection, we resume training the 4D Gaussians from the same $D_b$.
Since the Gaussians change slightly, the deformation network still possesses certain generalization ability, which helps reduce the time for training 4D Gaussians.

\vspace{-10pt}
\section{Conclusion}
\label{sec:conclu}
\vspace{-10pt}
In this paper, we introduce \name{}, a framework for comprehensive and complex 4D scene generation.
To address the limitations in previous 4D scene generation methods, we disentangle the generated reference video into foreground with rapid articulations and background subtle dynamics.
Then we generate them separately with our proposed parametric trajectory in the Dynamic Foreground Generation module and the novel progressive outpaint in our Dynamic Background Synthesis branch.
We finally compose the foreground and background to form the complicated 4D scene with depth information.
Through sufficient experiments and comparisons, we prove that our \name{} show superior 4D scene generation capability, representing a critical step towards the fully digital world in the future.

\clearpage
{\small
    \bibliographystyle{plain}
    \bibliography{neurips2025_conference}
}

\newpage
\appendix
\label{appendix}
\section*{Appendix}
\section{More implementation details}
\label{app:implementation}
In this section, we give a more detailed explanation of our implementation settings.
We use RTX6000 GPUs for all our experiments and ablations.
\begin{table}[h]
    \centering
    \caption{A detailed hyper-parameter list of our dynamic foreground generation part.}
    \begin{tabular}{l | c }
        \toprule
        Hyper-parameters & value \\ 
        \midrule
        \multicolumn{2}{c}{Gaussian learning parameters (static training)} \\
        \midrule
        initial points & 5000 \\
        feature\_lr & 0.01  \\
        position\_lr & 0.001  \\
        scaling\_lr & 0.005  \\
        opacity\_lr & 0.05 \\ 
        rotation\_lr & 0.005 \\ 
        densification\_interval & 100 \\ 
        total\_iter & 1000 \\
        \midrule
        \multicolumn{2}{c}{Deformation Network parameters (dynamic training)} \\
        \midrule
        deformation\_lr & 0.00064  \\
        grid\_lr & 0.0064  \\
        total\_iter & 800 \\
         \bottomrule
    \end{tabular}
    \label{tab:hyp_param_fg}
\end{table}
We first display the specific hyper-parameters of our training process for both foreground and background.
We list them in Tab.~\ref{tab:hyp_param_fg} and Tab.~\ref{tab:hyp_param_bg}.
\begin{table}[h]
    \centering
    \caption{A detailed hyper-parameter list of our dynamic background synthesis part.}
    \begin{tabular}{l | c }
        \toprule
        Hyper-parameters & value \\ 
        \midrule
        \multicolumn{2}{c}{Gaussian learning parameters} \\
        \midrule
        feature\_lr & 0.01  \\
        position\_lr & 0.001  \\
        scaling\_lr & 0.005  \\
        opacity\_lr & 0.05 \\ 
        rotation\_lr & 0.005 \\ 
        densification\_interval & 100 \\ 
        densification\_threshold & 0.5 \\
        max\_scaling & 0.2 \\
        total\_iter & 300 \\
        \midrule
        \multicolumn{2}{c}{Deformation Network parameters} \\
        \midrule
        deformation\_lr & 0.0064  \\
        grid\_lr & 0.064  \\
        total\_iter & 300 per camera pose \\
         \bottomrule
    \end{tabular}
    \label{tab:hyp_param_bg}
\end{table}.
For the trajectory learning, we simply give a learning rate of 0.1 to both the shift $T$ and scale $S$ during refinement.
In practice, we found that only 50 iterations of refinement can produce precise trajectory in the screen space.
What's more, we elaborate more on the experimental settings of our 3DGS projection and 4DGS training process.
Since a higher resolution of the reference video $V_{ref}$ in the 3DGS projection operation will cause a quadratic increase in the number of Gaussians, which will finally cause the excessive usage of GPU memory.
We thus first resize the reference video to a resolution of 256 * 256 to decrease the memory usage of our method.
After projection, we can render the 3DGS/4DGS at a higher resolution. With the help of RGB losses, we can train the background scene at the original resolution, which helps improve the fidelity and visual quality of the generated background scene.
We use Tab.~\ref{tab: memory} to illustrate how the resolution change will influence the memory cost.
\begin{table}[htp]
\centering
\renewcommand\arraystretch{1.}
\caption{Memory cost related to the change of projection resolution during training.}
\vspace{.1in}
\begin{tabular}{l | c}
\toprule
Resolution & memory consumption \\
\midrule
256 * 256 & 18 G\\
324 * 324 & 34.8 G\\ 
480 * 480 & 56.6 G\\ 
512 * 512 & Out of Memory (>80G)\\ 
\bottomrule
\end{tabular}
\label{tab: memory}
\vspace{-.1in}
\end{table}

Moreover, we show the settings of our progressive outpaint module in practical use.
We use a default radius around the coordinate center. For the camera intrinsics, we use an fovy of 60, a camera visible range of 0.1 to 10000.
We process two outpaint loops in practice. For the first loop, we have 4 camera poses in $C^1_{out}$, the elevation and azimuth of each camera pose are:(0°, 30°), (0°, -30°), (15°, 0°), (-15°, 0°), respectively.
For the second loop, we also have 4 camera poses in $C^2_{out}$, and the elevation and azimuth degrees of these camera poses are (15°, 30°), (15°, -30°), (-15°, 30°), and (-15°, -30°).
It is worth noting that to ensure stability in the inpainting process, we truncate the scaling of each Gaussian in the background scene to 0.2 (after processing the activation function) to prevent Gaussians from rendering protruding noises that will affect the rendered image and masks.

When composing the foreground object and background scene, we render the two different screen space pixels and compose them in the final camera presentation.

\section{Analysis on other 4D scene generation methods}
\label{app:analy}
\begin{table}[htp]
\centering
\renewcommand\arraystretch{1.}
\caption{Time cost for current text-to-4D scene generation methods.}
\vspace{.1in}
\begin{tabular}{l | c}
\toprule
Method & Time cost \\
\midrule
4D-fy& 24 hours\\
Dream-in-4D& 13.5 hours\\ 
VividDream& 3.5 hours\\ 
PaintScene4D& 3 hours\\ 
 Free4D&1 hour\\
 Ours&1 hour\\
\bottomrule
\end{tabular}
\label{tab: timecost}
\vspace{-.1in}
\end{table}

First, we present the time cost for all current text-to-4D scene generation methods in Fig.~\ref{tab: timecost}.
It is obvious that our method shortens the time cost for generation to 1 hour, which is the same as Free4D, and much shorter than both Score Distillation-based methods and Multi-view video training methods, which can be acceptable to users in practical generation.
What's more, since our method can enable immersive scene composition, it could be a better scheme for dynamic scene generation in applications like AR and VR.

\textbf{Discussion with other methods.} 
In the following, we analyze the advantages and drawbacks of the methods listed in Tab.~\ref{tab: timecost}. We select several representative methods for discussion and analysis.

\begin{figure}[htbp]
    \centering
\includegraphics[width=0.8\linewidth]{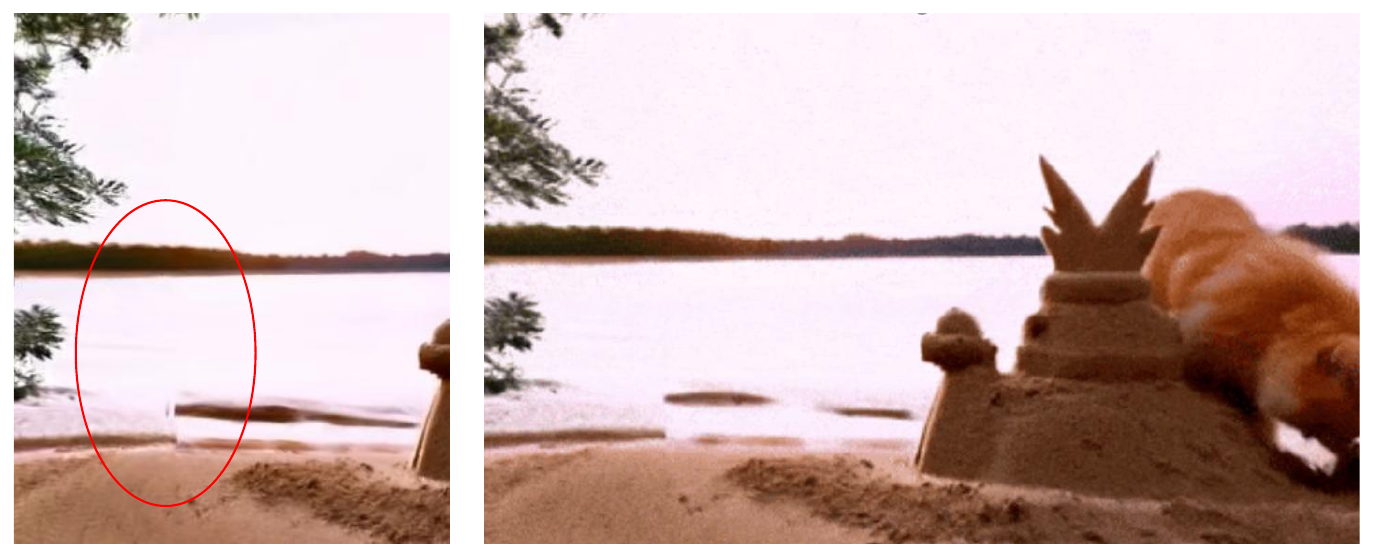}
    \caption{Drawback of PaintScene4D, caused by the inconsistent view-time matrix generated with image-inpainting.}
    \label{fig:paintscene4d}
\end{figure}

\textbf{4D-fy and Dream-in-4D.} 
These are two representative works of Score Distillation-based 4D generation methods.
And the most notable disadvantage is the generation time, which largely hinders the application of these methods and further research.
Another critical challenge in these works is the incorporation of dynamics.
Score distillation in the temporal dimension combines visual and temporal signals, but struggles to maintain latent space temporal consistency due to the sparse nature of distillation iterations.
In contrast, these methods are sophisticated in generating object-level 4D assets with a subtle dynamic degree.
But they still suffer from the Janus problem, which is inherited from the process of score distillation.

\textbf{PaintScene4D} generates the View-Time matrix with an image inpainting-based method, and as shown in their results in Fig.~\ref{fig:paintscene4d}, the view-time matrix tends to produce inconsistencies in the background regions with minimal motion patterns.
In opposite, our \name{} can seamlessly inpaint the novel views and produce consistent motion in the background region.
Fig.~\ref{fig:seperate} shows an example of the water wave in novel views.

\begin{figure}[h!]
    \centering
    \includegraphics[width=1.0\linewidth]{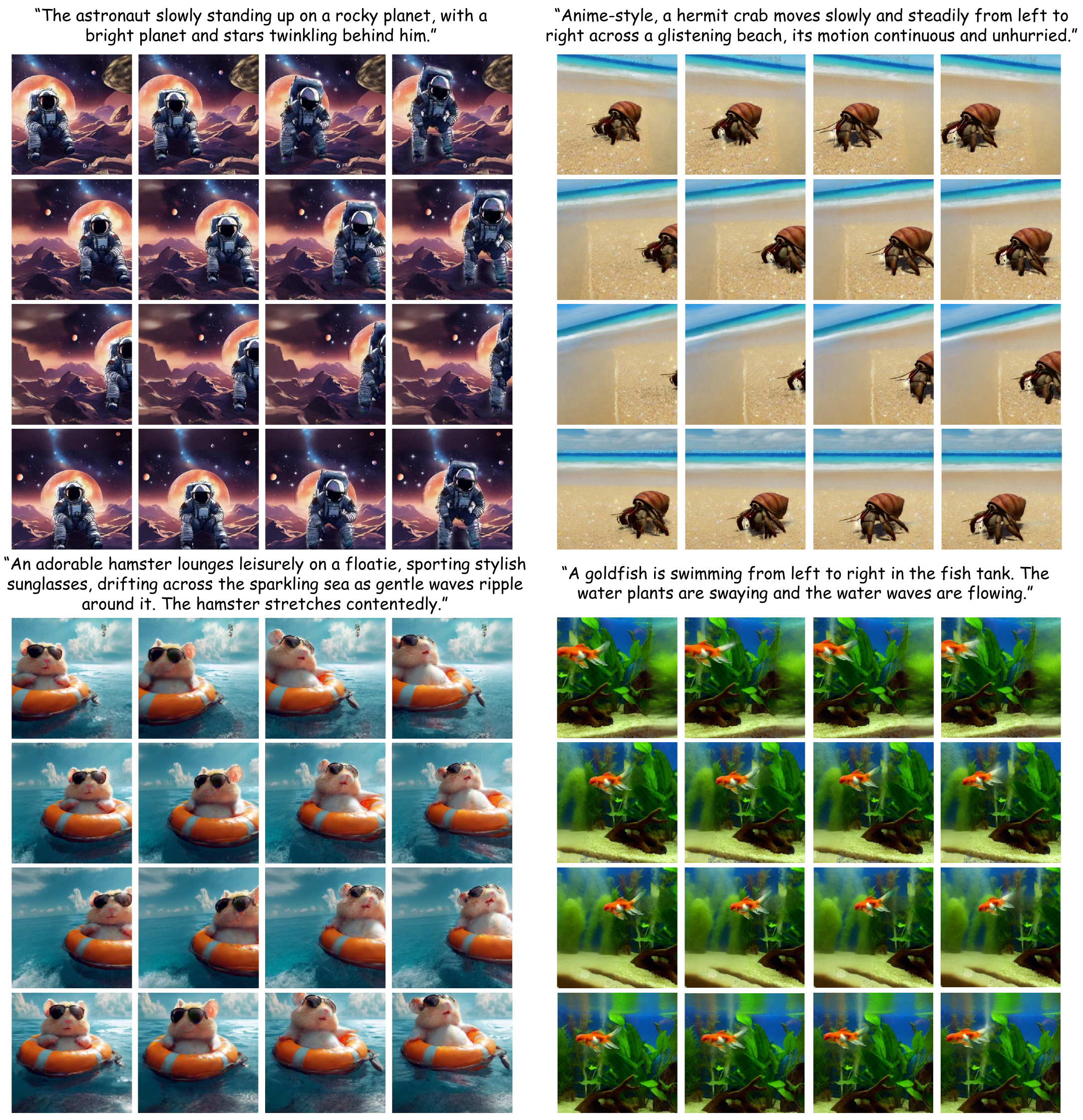}
    \vspace{-.2in}
    \caption{Additional visualization results of our \name{}.}
    \label{fig:add_vis}
    \vspace{-5pt}
\end{figure}

\textbf{Free4D} is the latest work on 4D scene generation.
It first uses a 3D-aware diffusion model~\cite{yu2024viewcrafter} to create point clouds that can serve as guidance for the completion of multi-view videos.
However, in our practical usage, Free4D often produces failures in the stereo initialization process. We attribute this issue to the excessively difficult novel-view synthesis process for video content with articulated motion and long trajectories.
What's more, even if we can initialize the stereo successfully, the multi-view training data is distorted because of the inherent drawbacks of point cloud representation: weak novel view synthesis capability.
Thus, when rendering at another camera pose, the existing part in the point cloud is shifted or even distorted.
A possible way to alleviate this might be to decrease the visible range of Free4D, even if it will hinder its application in real-world scenarios. 

\paragraph{Metrics.}
We discuss the metrics in this paragraph.
Since CLIP Score is kind of inaccurate and unreliable in assessing the quality of video frames, we supply it with some additional metrics in Vbench for more reliable comparison.
The selected three aspects are what we regard as most significant parts for a rendered video, which reflects the quality of the generated 4D scene best.

\section{Additional Visualization}
\label{app:vis}
We present additional visualization results in this section.
As shown in Fig.~\ref{fig:add_vis}, our \name{} generates complicated 4D scenes with a larger visible range and mature trajectory, which poses a critical challenge in previous 4D scene generation methods.
We can see from the visualization results that our method proposes consistent and high-quality results for both foreground objects and background scenes.
In the background scene, our \name{} produces consistent motion patterns (e.g., the water waving in the left lower example), compared to PaintScene4D and VividDream.

\section{Limitations}
\label{app:limitations}
Though \name{} can generate a comprehensive and complex 4D scene within an acceptable time cost, there are still some disadvantages.
First, the dynamic foreground generation will occasionally produce undesired results due to the bad results from Zero123.
Second, when projecting and optimizing the background scenes, inaccurate depth maps would cause artifacts in the final 4D scene, which would affect the quality of rendering.
Finally, the scene composition using depth information is not always precise, misalignment sometimes occurs, which damages the spatial structure of the 4D scene.

\end{document}